\documentclass{article}

\usepackage{arxiv}

\usepackage[utf8]{inputenc} % allow utf-8 input
\usepackage[T1]{fontenc}    % use 8-bit T1 fonts
\usepackage{hyperref}       % hyperlinks
\usepackage{url}            % simple URL typesetting
\usepackage{booktabs}       % professional-quality tables
\usepackage{amsfonts}       % blackboard math symbols
\usepackage{nicefrac}       % compact symbols for 1/2, etc.
\usepackage{microtype}      % microtypography
\usepackage{lipsum}
\usepackage{graphicx}
\usepackage{microtype}
\usepackage{graphicx}
\usepackage{subfigure}
\usepackage{booktabs} % for professional tables
\usepackage{float}
\usepackage{ifthen}
\usepackage[numbers]{natbib}
\usepackage{doi}
\usepackage{amsmath}

\usepackage{hyperref}
\hypersetup{colorlinks=true, urlcolor=blue}

\title{Scalable federated machine learning with FEDn}

\author{Morgan Ekmefjord\\
        Scaleout Systems\\
        Uppsala, Sweden\\
        \texttt{morgan@scaleoutsystems.com}
        \And
        Addi Ait-Mlouk\\
        Department of Information Technology\\
        Uppsala University, Uppsala, Sweden\\
        \texttt{addi.ait-mlouk@it.uu.se}
         \And
        Sadi Alawadi\\
        Department of Information Technology\\
        Uppsala University, Uppsala, Sweden\\
        \texttt{sadi.alawadi@it.uu.se}
         \And
        Mattias {Åkesson}\\
        Scaleout Systems\\
        Uppsala, Sweden\\
        \texttt{mattias@scaleoutsystems.com}
         \And
        Prashant Singh\\
        Department of Information Technology\\
        Uppsala University, Uppsala, Sweden\\
        \texttt{prashant.singh@it.uu.se}
         \And
        Ola Spjuth\\
        Department of Pharmaceutical Biosciences\\
        Uppsala University, Uppsala, Sweden\\
        Scaleout Systems, Uppsala, Sweden\\
        \texttt{ola.spjuth@farmbio.uu.se}
        \And
        Salman Toor\\
        Department of Information Technology\\
        Uppsala University, Uppsala, Sweden\\
        Scaleout Systems, Uppsala, Sweden\\
        \texttt{salman.toor@it.uu.se}
         \And
         Andreas Hellander\\
        Department of Information Technology\\
        Uppsala University, Uppsala, Sweden\\
        Scaleout Systems, Uppsala, Sweden\\
        \texttt{andreas.hellander@it.uu.se \thanks{Corresponding author.}}
}

\renewcommand{\shorttitle}{Scalable federated machine learning with FEDn}

\hypersetup{
pdftitle={Scalable federated machine learning with FEDn},
pdfsubject={q-bio.NC, q-bio.QM},
pdfauthor={Morgan Ekmefjord,Addi Ait-Mlouk,Sadi Alawadi, Mattias Åkesson, Desislava Stoyanova,Ola Spjuth,Salman Toor,Andreas Hellander},
pdfkeywords={Federated machine learning , artificial intelligence,privacy},
}

\begin{document}
\maketitle
\begin{abstract}
Federated machine learning promises to overcome the input privacy challenge in machine learning. By iteratively updating a model on private clients and aggregating these local model updates into a global federated model, private data is incorporated in the federated model without needing to share and expose that data. Several open software projects for federated learning have appeared. Most of them focuses on supporting flexible experimentation with different model aggregation schemes and with different privacy-enhancing technologies. However, there is a lack of open frameworks that focuses on critical distributed computing aspects of the problem such as scalability and resilience. It is a big step to take for a data scientist to go from an experimental sandbox to testing their federated schemes at scale in real-world geographically distributed settings. To bridge this gap we have designed and developed a production-grade hierarchical federated learning framework, FEDn. The framework is specifically designed to make it easy to go from local development in pseudo-distributed mode to horizontally scalable distributed deployments. FEDn both aims to be production grade for industrial applications and a flexible research tool to explore real-world performance of novel federated algorithms and the framework has been used in number of industrial and academic R\&D projects. In this paper we present the architecture and implementation of FEDn. We demonstrate the framework's scalability and efficiency in evaluations based on two case-studies representative for a cross-silo and a cross-device use-case respectively.
\end{abstract}


\section{Introduction}

Federated machine learning (FL) \cite{mcmahan2017communication,kairouz2019advances} is a promising solution to the input data privacy problem in machine learning. In FL, multiple parties, or clients, jointly train a machine learning model while keeping all training data local and private. Instead of moving data to a central storage system, 
computation is brought to data at each local client site, and incremental model updates are computed and then combined into a global model according to some aggregation scheme. In this way, only model parameters are shared. The majority of work has dealt with artificial neural networks  \cite{mcmahan2017communication,bacon,bonawitz2019towards,sheller2018multi} but there is also research done on federated versions of other statistical learning methods such as random forests \cite{liu2019revocable}.

FL is structurally similar to distributed optimization for statistical learning \cite{boyd2011distributed,xing2016strategies,186214} but differs in a number of important ways. Foremost, in FL there is no control over how the data is distributed over the participating computational nodes. Hence, data cannot be assumed to be balanced or IID across nodes. This is in stark contrast to distributed learning where we are in full control of the data partitioning and are this able to devise optimal partitions for convergence and load balancing. Moreover, whereas distributed learning typically occurs on reliable cluster infrastructure with high network performance and low failure rate, in FL clients can become unavailable at any time, and communication between client and server takes place over a high-latency network (the internet). Consequently, how to optimally balance local training iterations with global synchronization to avoid poor convergence and to minimize communication rounds are central research questions \cite{bonawitz2019towards,li2019convergence,kairouz2019advances}. Benchmark suits targeted at the exploration of federated learning algorithms have also been proposed \cite{caldas2018leaf}. 

Real-world federated learning software needs to be robust, resilient, and highly scalable distributed systems capable of handling both scaling to a large number of edge-clients, and scaling to large model sizes. The design and implementation of such a framework is the main contribution of this paper. Due to a tiered architecture inspired by the MapReduce programming model, FEDn users can implement a wide range of federated learning schemes by adhering to a structured and familiar design pattern, with the framework then assuring that the schemes will be highly resilient and horizontally scalable. FEDn aims to be easy-to-use and agnostic when it comes to the ML-framework used by clients while supporting high-performance federated training in real distributed settings. It is thus well-suited as a tool for research that bridges the machine learning aspects of the problem and the systems aspect of the problem, as well as for running federated learning deployments in production. 

FEDn has been used to implement a number of federated learning projects including a Federated Electra by the Swedish National Library, Federated Object Detection for Baltic seabirds by AI Sweden and Zenseact researchers, and fully distributed deployments of FEDn is available for wide use by researchers and Swedish industry partners in the strategic edge computing testbed AI Sweden EdgeLab. Links to these examples and others are collected and at the Scaleout organization on GitHub. FEDn is also the core enabling framework for federated machine learning in the SESAR project AICHAIN. 

We make the following main contributions: 

\begin{itemize}
    \item We propose a tiered system architecture based on hierarchical FL and inspired by the MapReduce programming model. In this way we provide a programming pattern for FL applications that ensures highly scalable and resilient federated learning for cross-silo and cross-device scenarios. 
    \item We provide a highly efficient open source framework, FEDn, implementing the proposed architecture. FEDn lets a user go from local testing and development to production-grade geographically distributed deployments with no code change.  
    \item We provide performance benchmarks based on thousands (cross-device) of geographically distributed clients and for machine learning model sizes ranging from a few kB to 1GB (cross-silo). Systematic and geographically distributed benchmarks of live federated training at this scale has, to the best of our knowledge, not been reported previously.  
\end{itemize}

The remainder of this paper is organized as follows. Section~\ref{sec:background} gives a background and surveys related work. Section~\ref{sec:approach} explains the architecture and implementation of FEDn. In~Section~\ref{sec:experiments} we demonstrate the performance and scalability of the framework through experiments on both cross-silo and cross-device use-cases. Finally, Section~\ref{sec:conclusion} concludes the work and outlines future~directions.

\section{Background} \label{sec:background}
\subsection{Horizontal and vertical federated learning} 
It is common to divide federated learning cases in two distinct categories \cite{yang2019federated}. In \emph{horizontal FL} the data is assumed to be partitioned by example, i.e. the feature space are assumed to be (nearly) the same between different participants (for example different scientific instruments sharing performance metrics but operating in different conditions and owned by different companies). In \emph{vertical FL} the data is assumed to be partitioned by features, i.e. different participants might hold different types of information about the same example (for example a bank and an insurance company holding different types of data about the same customer). The present study focuses on horizontal FL applications. The majority of FL solutions in literature fall in the horizontal FL category, often based on Artificial Neural Networks (ANNs), and perform synchronous batch-based training rounds, see e.g. \cite{wu2020privacy,smith2018don}. Recently, Bonawitz et al. \cite{bonawitz2019towards} proposed a high-level design of a scalable cross-device framework based on TensorFlow where the focus is on horizontal FL training. Kewei et al. \cite{DBLP:journals/corr/abs-1901-08755} introduced a lossless federated training scheme based on secure XGBoost for vertical FL. 

Federated averaging (FedAvg) \cite{mcmahan2017communication} is the most widely used method for horizontal FL. It is a decentralized version of stochastic gradient descent (or in general any method that relies on a gradient update scheme) where in one \emph{round} of training, a subset of $M$ participating clients receive a copy of the latest global model and execute $K$ local epochs ( complete passes over data) of training, updating $f(w_k)$ locally on their own private datasets $D_k$. They then send the updated weights $w_k$ to the server which averages those weights:

 \begin{align}
 w^{(i)} = \sum_{k=1}^{M} \frac{n_k}{n}w_k^{(i)}.  
 \end{align}

These averaged weights are then communicated to all clients and this concludes one \emph{round} of training. The currently supported scheme in FEDn is based on a hierarchical implementation of FedAvg. The choice of hyperparameters, in particular the number of local epochs $K$, critically influences the convergence rate and overall training efficiency - a too large value risks local overfitting and slow convergence of the global model, while a small value increases the number of communication rounds. Optimal values will depend both on the model, the number of clients and the data distribution over clients. In a recent study, Pang et al. \cite{pang2020realizing} investigated the impact of the data heterogeneity problem in FL. Another study proposed a self-balancing federated learning framework Astraea \cite{duan2019astraea} to deal with imbalanced clients' data. Lim et al. \cite{lim2020federated} presented a number of different strategies to address communication overhead and unreliable network speed. In this paper we do not aim to select optimal hyperparameters for our use-cases, but rather focus on the cost of executing each round as a function of number of clients and model size (system performance rather than model performance).      

\subsection{Cross-device vs cross-silo use cases}
We can distinguish between two major settings based on the nature of the use-case and participant/client type \cite{kairouz2019advances}. In \emph{cross device learning} the target is typically a very large amount of stateless, low-powered clients engaging in relatively cheap, short bursts of computation. Clients can for example be cell phones, IoT devices or vehicles, and data is partitioned horizontally.  
For \emph{cross-silo} use-cases, clients are larger entities with more local computational power and storage capability. Data can be partitioned horizontally or vertically. In cross-device learning, communication overhead and handling failed connections are key challenges, whereas cross-silo learning can be both computationally and communication constrained and models are relatively large (MBs to GBs). FEDn aims to support the requirements across both these axes of federated learning though a horizontally scalable architecture. 

\subsection{Privacy-enhancing technologies}
The core contribution of FL in the wider context of privacy-preserving machine learning is to enhance input privacy by allowing training data to remain withing the full control of the data owner. Compared to alternatives such as homomorphic encryption (HE) which allows computation directly on encrypted data \cite{halevi2014algorithms}, FL accommodates a wider range of algorithms. \emph{Output privacy} deals with the inverse problem of what can be learnt about the input data from making predictions with the resulting global model. Differential privacy (DP) can be used to add controlled noise to carefully chosen stages of the model construction pipeline in order to make it harder to reverse engineer input data \cite{abadi2016deep}. It is common to combine FL with HE and DP \cite{ryffel2018generic}, at least with synchronous model update primitives \cite{kairouz2019advances}. Multiparty computation (MPC) can also be used for secure aggregation of models to minimize risk for information leakage \cite{bonawitz2019towards}. 

A comprehensive overview of the current state-of-the art in FL along with a survey of privacy-enhancing technologies and security is provided by Kairouz et al.  \cite{kairouz2019advances}. 

In this paper we focus on efficient implementation of the core primitives of FL, but we note that privacy-enhancing technologies, in particular differential privacy and multiparty computation, could be accommodated in the framework with no or limited architectural modifications.  

\subsection{Related work}
Several open-source frameworks for FL have been developed, including the following ones \cite{lim2020federated}.

\begin{enumerate}
    \item TensorFlow Federated (TFF).\footnote{\url{https://github.com/tensorflow/federated}} TFF comprises two layers: Federated Learning and Federated Core. The former is a high-level interface that supports users to apply FL without the need for implementing FL algorithms personally. The latter enables users to implement and experiment new or customized FL algorithms. TFF supports the simulation of the distributed training of FL models but executes only on a single machine.    
    
    \item PySyft\footnote{\url{https://github.com/OpenMined/PySyft}} leverages deep learning, secure multiparty computation, and differential privacy to train privacy-preserving FL models in untrusted environments \cite{ryffel2018generic}. The framework is based on PyTorch and provide a native Torch interface. 

    \item FATE \footnote{\url{https://github.com/FederatedAI/FATE}} is an open-source framework developed by Webank \cite{fate}. It exploits secure computation protocols to train FL models using different algorithms, such as decision trees, logistic regression, transfer learning, and deep learning.
    
   \item FedML \footnote{\url{https://fedml.ai/}} is an open-source library developed to facilitate the development and benchmarking of FL algorithms. FedML supports on-device training, distributed computing, and single-machine simulations. Further, it provides generic API design and reference baseline implementations \cite{he2020fedml}.
    
    \item PaddleFL \footnote{\url{https://github.com/PaddlePaddle/PaddleFL}} is an open-source framework that supports the replication and comparison of FL algorithms as well as the deployment of FL systems in distributed clusters.  
    
    \item TiFL is a Tier-based Federated Learning System that groups clients into tiers based on their training performance to mitigate the straggler problem caused by the heterogeneity of clients' capabilities or data quantity. 
    \cite{chai2020tifl}. 
    
     \item FLOWER\footnote{\url{https://github.com/adap/flower}} is a new open-source framework-agnostic by design that promotes various aggregation algorithms and deep learning frameworks (e.g. Tensorflow, MXNet,TFLite and PyTorch). Moreover, Flower supports training and evaluation on heterogeneous real-edge devices and multi-cluster nodes. 
     \cite{beutel2020flower}.
\end{enumerate}

In addition to the frameworks mentioned above, OpenFL proposed by Intel specializes in healthcare use-cases \cite{reina2021openfl}. NVIDIA has recently open sourced a standalone python library called NVFlare \cite{nvflare}.

 Unlike the above mentioned frameworks, the main aim of FEDn is to provide a production-grade and framework-agnostic distributed implementation with strong scalability and resilience features supporting both cross-silo and cross-device scenarios. To this effect, FEDn implements a two-tiered hierarchical federated learning architecture with a framework based loosely on the well-known MapReduce programming pattern. The work most closely related to ours is the architecture proposed in \cite{bonawitz2019towards}, where the authors seek to provide load-balancing capabilities through replication of local servers (with a similar role as combiners in our terminology, see the following sections). However, to the best of our knowledge no open source implementation associated with the work is available, and our work goes beyond that work in the size and scale for large model updates, and in terms of resiliency features.     
 
\section{A flexible and horizontally scalable architecture for federated learning}\label{sec:approach}

\subsection{Architecture overview}

\begin{figure}[htp!]
    \centering
    \includegraphics[width=\linewidth]{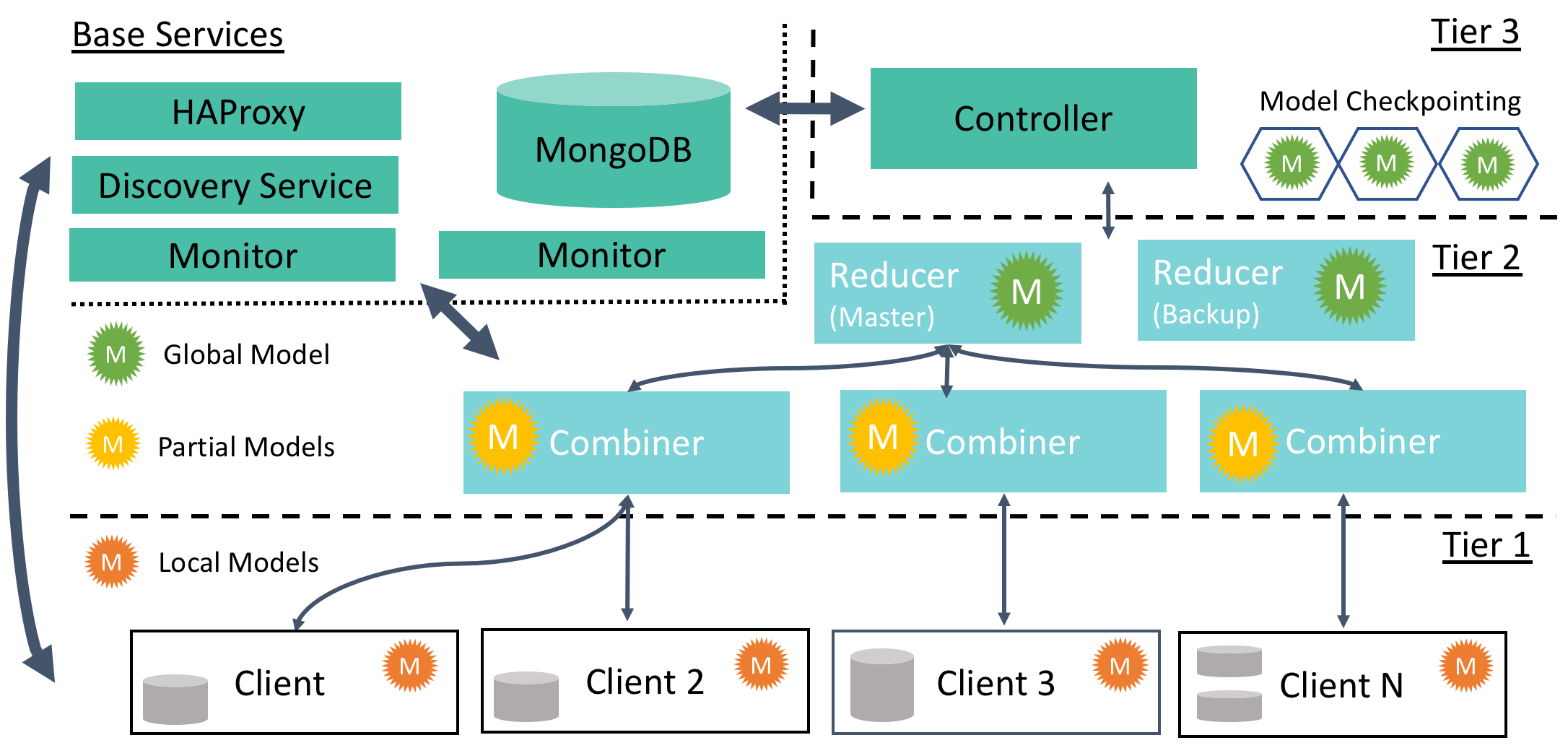}

    \caption{Overview of the computational model and logical architecture of FEDn. FL model training is organized in three logical layers, each populated by components with clearly defined roles. By scaling the number of Combiners we are able to meet communication demands imposed by a growing number of clients.}

    \label{fig:overview}
\end{figure}

Fig. \ref{fig:overview} illustrates the logical architecture of FEDn. Inspired by the map-reduce paradigm, a well-known scalable distributed systems design, the system consists of three logical tiers. The first tier consists of geographically distributed \emph{clients}. A client is responsible for local model updates and interfaces with the local data source. Clients are the only entity that touches the private datasets. 

The second tier is composed of one or many \emph{Combiners} and a \emph{Reducer}. Combiners are stateless servers with responsibility for coordinating updates from their own subset of clients. Together, the combiners and reducer make up the \emph{FEDn network}. The number of combiners needed in such a network depends on the number of clients and the size of models. Each combiner is responsible for producing a partial model update in a global round of federated training. The partial models are aggregations of model updates by the combiner's associated clients. 

The key components of the third tier is the \emph{Controller} and the \emph{Discovery service}. The controller is responsible for coordinating the overall computation done in a global training round, for maintaining a trail of global models (model checkpointing), and to handle global state. The Discovery service component is responsible for receiving client connection requests and assigning clients to combiners in the network. 

More details on the architecture and implementation can be found in the the online documentation at https://scaleoutsystems.github.io/fedn/.

\subsection{Design pattern for implementing FL schemes in FEDn}

All algorithms supported by FEDn should be able to scale horizontally to meet increasing load on the aggregation servers. For this reason, a user wanting to implement a custom FL scheme in FEDn does this by a) implementing the combiner lever aggregate function (aggregate partial models), b) implementing the Reducer level aggregation function (reduce combiner level models into a global model) and c) implement the global round controller. For each of these, abstract classes define the interface. The default scheme in FEDn employs FedAvg on both the combiner and reducer level resuling in hierarchical FedAvg. While this model may seem restrictive, in practice it enables a large space of possible schemes (for example by simply grouping combiner models on the reducer level, a wide range of ensemble strategies could be achieved).    

\subsection{Implementation of the FEDn network}

\begin{figure}[htp!]
    \centering
    \includegraphics[width=0.85\linewidth]{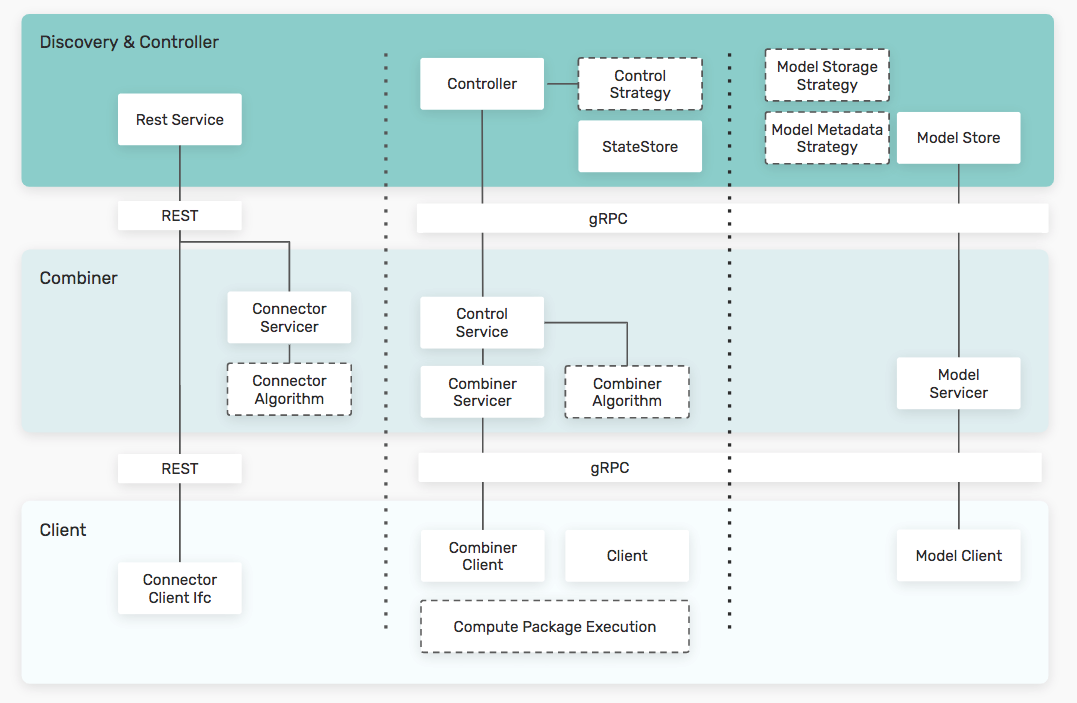}
    \caption{Implementation of FEDn. FEDn provides functionality for status logging, monitoring and visualization of training progress. The current implementation uses MongoDB to store logs and state, and S3/Minio to store the global model trail, however the API allows for different storage backends to be added.  
}
    \label{fig:comms}
\end{figure}

Fig. \ref{fig:comms} shows a schematic of the key components of the software. The central communication protocol between clients and combiners is built using Google Protocol buffers \cite{GPB} and gRPC \cite{gRPC}, a popular framework for remote procedure call (RPC). The main reason for this technology choice, apart from high-performance, is that it allows for a great deal of flexibility in the language choice when it comes to implementation of clients. This is important since clients may need to run in resource constrained systems, or on edge devices with special hardware and software environments. 
 
\subsubsection{Combiner} The Combiner (CB) is a stateless gRPC server whose main role is to coordinate client updates and to aggregate model updates from a given subset of clients. During a model update round, each combiner operates completely independently of all other combiners in the network. At any given point in time, the combiner works on one and only one partial model update. The fact that the combiner is stateless makes the network highly fault-tolerant -- a combiner that becomes unresponsive can easily be replaced, and in the processes only model updates from its associated clients for that particular round is lost. It also results in a horizontally scalable network since there is no communication between combiners. The work done by the combiner scales linearly with the number of attached clients, and the total number of clients that can be supported by the network is directly proportional to the number of combiners. 

\subsubsection{Reducer} The reducer is responsible for implementing and executing a reducer protocol, combining all partial model updates computed by combiners into a single global model update in each round. FEDn supports use of multiple reducers running in  active-passive mode. At any given time, one reduce is responsible for preparing the global model. In the current implementation, the reduce operation is implemented as a separate service that pulls partial model updates from the combiners via a gRPC stream and incrementally aggregates these updates. However, in general the reducer protocol can be extended to execute e.g. hierarchical reduce directly using combiners, or with a secure multiparty computation protocol (but this would affect scalability properties). The work done by the reducer is scales with the number of combiners (in a protocol-dependent manner, linearly in the current implementation) and is independent of the number of clients.

\subsubsection{Clients}
A client (CL) is the main worker in the system. It is an actor that runs on a local site or device and is able to directly access local private data. Clients join a network by asking the discovery service for a combiner assignment. It then connects to its combiner and receives training and validation requests, downloads the global model from the combiner, executes model updates and validations, and streams the results back to the combiner. At the initialization of a federated model, a \emph{compute package} is prepared by the model initiator and uploaded to the Controller. When a client sends a connection request to the discovery service, it receives this compute package from the controller and stages it in its local execution environment. Two design objectives related to clients have been central when developing FEDn: 1) \emph{ML framework and language agnostic.} To support a wide range of applications, we use a black-box execution model. Each invocation of either a model update or model validation task executes as a single-input single-output (SISO) program. The only requirement from the framework's perspective is that it knows how to serialize and deserialize the input and output. This allows for a ML-framework agnostic implementation and we currently support both Keras/Tensorflow and PyTorch out of the box (examples using both frameworks are provided in the Supplementary Manuscript). It also means that we cannot use any knowledge about the internals of the network for optimizations or fine-grained parallelism since the unit of computation is the complete model.
2) \emph{No incoming connections allowed.} It should not be necessary to have any open ingress ports on the client host. This is a firm requirement in many production scenarios. A client is an gRPC client (Fig. \ref{fig:comms}C), which listens for messages on a unidirectional streaming RPC service endpoint on the combiner. Model updates are downloaded and uploaded to the combiner using a dedicated gRPC Model service (Fig. \ref{fig:comms}D). These services are responsible for chunking the binary large objects (model updates) and for data transfer using server streaming RPC and client streaming RPC respectively.

\subsection{Controller Round Protocol}
The controller implements a configurable round protocol. During each global round of training, the following sequence of events take place in the default implementation: 
\begin{enumerate}
    \item Ask all combiners if they can participate in the round using the \emph{round participation policy}. 
    \item Set a deadline and ask all participating combiners to coordinate a partial model update.
    \item Wait until all combiners report a completed update, or the round times out. 
    \item Check if the round should be considered valid by evaluating a \emph{round validity policy}.
    \item If the round is deemed valid, ask the reducer to aggregate all combiner partial models into the global model. 
    \item Ask combiners to coordinate \emph{model validations} (optional). 
    \item Commit the global model to the model trail. 
\end{enumerate}
Global rounds are repeated as needed for model convergence. Several steps in this processes can be configured, hence altering the detailed behavior in a round. For example, the round validity policy can be used to enforce certain requirements on how many combiners need to be successful in a round for the global model to be updated (the default is one). 

\subsection{Resilience}
The training process is able to recover robustly in the events of failures in the FEDn network. Resilience is vital for production grade federated machine learning. The strategy in FEDn is horizontal component replication at each tier. 

The clients in the first tier can join and leave the framework at any time. If a client drops out during a round, its controlling combiner waits for a configurable amount of time (model update deadline), then proceeds with its next set of instructions. A client can simply rejoin at a later time. In practice the deadline is a parameter that should be chosen according to the training cost and balance risk of delays vs risk of missing model updates in a round. Future work will provide additional guidelines and support for choosing this system parameter.  

For each client, discovery of combiners is dynamic. The available discovery service assign a combiner to each client. If a combiner becomes unavailable or have some issues during run time, clients get automatically reassigned to another available combiner.   

The reducer component is responsible for the generation of the global model. In classical federated machine learning settings, a reducer can be viewed as a single point-of-failure. However, the FEDn architecture offers a horizontally scalable reducer setup with active-passive settings. At any given time, one reducer is responsible for the global model preparation to ensure consistency. In case of network disruption or other failures, a passive reducer becomes active and offer services. Here it is important to note that if a failure occurs during an active training round, the training process stops to avoid the risk of inconsistency in the models. The user can simply resume training from the last completed round. Supplementary videos demonstrating the resiliency of the framework can be found on the Scaleout YouTube channel \footnote{\url{https://www.youtube.com/channel/UCZVv30LFXMQUOswNDKuQpNA}}. 

Finally, all base services needed by FEDn also follow the same strategy of replication. The backend database, MongoDB can be deployed with high availability settings. Other services, discovery and monitors are all stateless units that can scale horizontally. All the replicated units can have a signal entry point using HAProxy \cite{haproxy}.    

\section{Evaluation} \label{sec:experiments}
To evaluate the performance and scalability of FEDn, we consider two use-cases: 1) a cross-silo natural language processing use-case and 2) a cross-device use-case based on digital human activity recognition. In the cross-silo case, we focus on the cost of a global round of federated training as a function of model size focusing on horizontal scalability and throughput in a geographically distributed setting. In the cross-device case, we benchmark single combiner performance with lightweight clients in the range of 100-1000 clients demonstrating framework implementation efficiency focusing on resource utilization. Taking together, our experiments highlight that the proposed architecture and implementation performs well for both these application classes. We stress that the goal of our evaluations is not to study the accuracy of the federated models as such, but rather to investigate how the cost of each global round depends on the problem class characteristic. With that said, it is important to ensure the correctness of the implementation. A convergence study and comparisons to a centrally trained model for the common CIFAR-10 benchmark dataset with a VGG-16 model and a PyTorch client is available at: https://github.com/scaleoutsystems/FEDn-client-cifar10-pytorch.  
      
\subsection{Horizontal scalability for a geographically distributed cross-silo use-case} 
A central task in natural language processing (NLP) is the generation of word embeddings (representation models), i.e. representations of every word in a vector space, allowing words with similar meaning to have a similar representation. 
Large scale representation models such as Word2Vec \cite{NIPS2013_5021}, Glove \cite{Pennington14glove:global}, fasText \cite{bojanowski-etal-2017-enriching}, Elmo \cite{Peters:2018}, and BERT \cite{devlin-etal-2019-bert}, have significantly advanced NLP in recent years.

Private companies or government agencies (e.g., banks, hospitals, and institutes), each own unique large corpora with specific information. These corpora might not be complete enough to individually train high-quality NLP tasks. These organizations may want to collaborate to build a unified model, but without sharing the contents of their corpora. 
FL thus holds large potential to advance NLP. However, the size of such models poses a challenge for FL systems due to the large data transfer requirements.

\begin{table}[htp]
% \scriptsize
\caption{Model update size and number of parameters}
\label{nlp-params}
\begin{center}
\begin{small}
\begin{sc}
\scalebox{.85}{ 
\begin{tabular}{ccccc}
\hline
10 MB & 50MB & 100MB & 200MB & 1GB \\
\hline
2638895 & 13624335& 27124335 & 54664335 & 273329135\\
\hline
\end{tabular}
}
\end{sc}
\end{small}
\end{center}
\end{table}

The purpose of this example is to demonstrate how the FEDn architecture enables large cross-silo use-cases by scaling the combiner network. 
We conducted experiments using the IMDB dataset with different model sizes (10MB, 50MB, 100MB, 200MB, 1GB). The model is generated using: 1) Embedding layer (max\_features = 20000, maxlen = 100, embedding\_size = 128), 2) Convolution (kernel\_size = 5,  filters = 64, pool\_size = 4), 3) LSTM (lstm\_output\_size = 70) and the local training parameters per round are: 1 epoch, batch size 32 and ADAM as the optimizer. The different model sizes were generated by tuning the Embedding layer of the seed model, see the Table \ref{nlp-params}. For all experiments we used cloud computing resources from two clouds; SNIC Science Cloud, a Swedish National OpenStack cloud, and Amazon Elastic Compute Cloud (AWS EC2). All VMs are based on Ubuntu 20.04 Server LTS.   

\subsubsection{Validation of FEDn training accuracy}

As mentioned earlier, factors affecting convergence rate and thus the number of global rounds is an active research area out of the scope of this study. However, to demonstrate both the correctness of the FEDn implementation, we partitioned the dataset in 10 equal chunks, so that each client has $10\%$ of the total dataset. We then compare four federated scenarios to centralized model training: 1) 2 clients for 50 rounds (1CB-2CL) 2),  5 clients (1CB-5CL) and 3) 10 clients (1CB-10CL).

\begin{figure}[H]
    \centering
    \includegraphics[width=\linewidth]{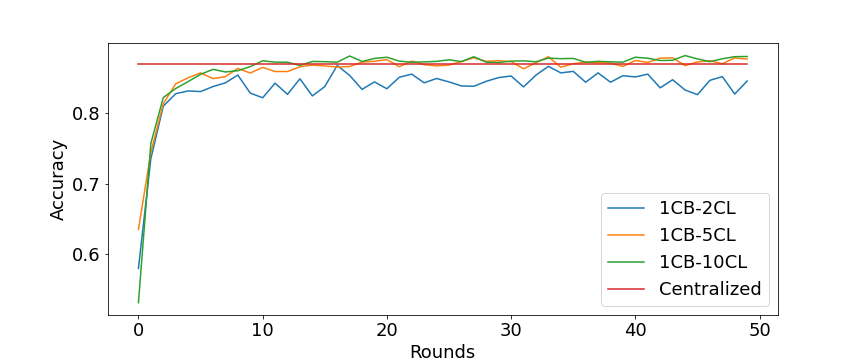}
    \caption{Convergence comparison on IMDB dataset in three different scenarios: 1) 1 combiner, 2 clients and 50 rounds 2) 1 combiner, 5 clients and 50 rounds, 3) 1 combiner, 10 clients and 50 rounds.}
    \label{fig:accuracy}
\end{figure}

The result in Fig. \ref{fig:accuracy} shows that FL in FEDn reaches the baseline centralized performance (horizontal line) as more clients (and hence more of the data) are used. In particular, federated learning achieves 87$\%$ accuracy for 10 clients (baseline accuracy 86$\%$ in this experiment).

\subsubsection{The combiner round time scales linearly with model size}
\begin{figure}[htp]
    \centering
    \includegraphics[width=0.7\linewidth]{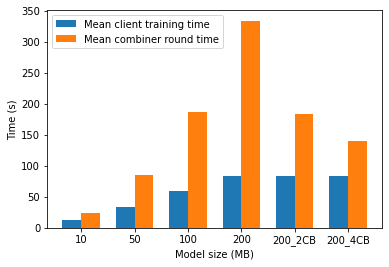}
    \caption{Combiner round times for increasingly large models (NLP downstream tasks).}
    \label{fig:round_time}
\end{figure}

We first consider a FEDn network consisting of a single, high-powered combiner (8 core, 16GB RAM) in SSC with 48 connected clients, 12 8VCPU,16GB instances in SSC and 12\emph{c5.2xlarge} instances in EC2 in the \emph{eu-north} region and measure the average round time over 5 global rounds. Figure \ref{fig:round_time} shows how the round time at the combiner is affected by increasing model size (orange bars). Since the model size affects both the training time at clients and the cost for data transfer and model aggregation, we show the mean training time for reference (blue bars).  

As can be seen, for the smallest model (10MB) the round time is marginally higher than the client training time and the single combiner can comfortably support the 48 clients. As we scale up the model size, as expected we see an almost linear increase in round time due to the increased communication cost of transferring updates. At 200MB updates the overhead is almost 300s per round. To demonstrate the load balancing capabilities, we next deployed additional combiners for a total of 2 combiners (\emph{200\_2CB}) and 4 combiners (\emph{200\_4CB}). As can be seen, this scaling of the FEDn network is an efficient way to bring round time down by balancing work over more aggregation servers.   
\begin{figure*}[htp]
  \centering
  \includegraphics[width=\linewidth, height=5.5cm]{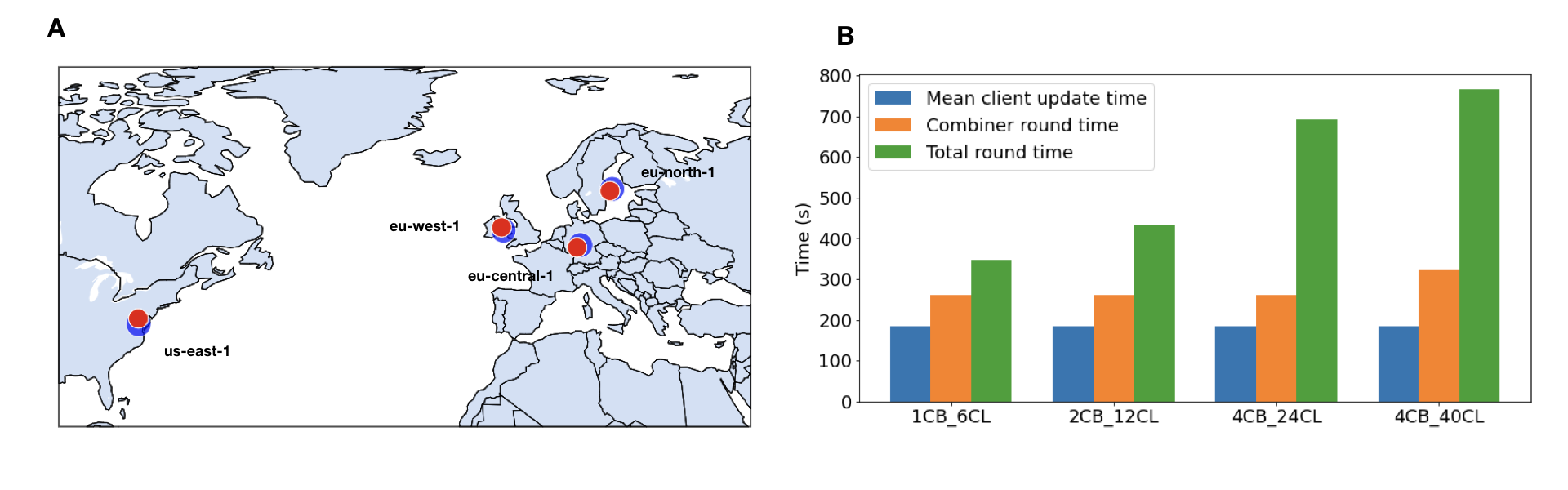}
  \caption{Geographically distributed horizontal scaling with 1GB model updates. Combiners are spread over 4 regions in AWS EC2 (A). This FEDn network accommodates 40 clients in this experiment, coordinating model updates of in total 40GB in each round (B).}
  \label{fig:fed-cross-silo}
\end{figure*}
To conclude, Fig. \ref{fig:round_time} shows that the mean combiner round time is to a large extent governed by the size of the model updates for a fixed number of clients, and that scaling the combiner network horizontally is an efficient way to reduce round time in this scenario.

\subsubsection{Multi-region horizontal scaling for large model updates}

Next we conducted a large-scale geographically distributed experiment using the 1GB version of the model. We deployed a FEDn network with in total 4 combiners, each on a large server (c5.4xlarge, 16VCPU,32GB RAM flavor) in four regions of AWS EC2, eu-north-1 (Stockholm), eu-west-1 (Ireland), us-east-1 (N. Virgina) and eu-central-1 (Frankfurt). The combiners are illustrated with blue dots in Fig. \ref{fig:fed-cross-silo}A and clients with red dots (one dot per region, many clients per dot). The reducer is deployed on a c5.4xlarge instance in the Stockholm region. 

We then conducted a weak scaling experiment, doubling both the number of clients and the number of combiners in three successive experiments by attaching 6 c5.2xlarge clients to each combiner, and measured the round time, see Fig. \ref{fig:fed-cross-silo}B 1CB\_6CL, 2CB\_12CL and 4CB\_24CL. As can be seen, the system is capable of accommodating 1GB model update transfers with growing number of clients at a near constant combiner network round time. However, as expected, the cost of the reducer, and hence the total round time, grows with the number of combiners. With this network configuration, a 4x increase in the number of clients led to a 2.0x increase in round time. This is in itself a good property, but also highlights an interesting aspect of sizing a FEDn network. In the final experiment we kept all 4 combiners fixed and allowed 16 more clients, 5 from eu-west-1 and 11 from eu-north-1, for a total of 40 clients, to connect to the combiners randomly, i.e. there is no guarantee that they will connect to a geographically close combiner (4CB\_40CL). As can be seen, there was capacity in the combiner network, and while the mean round time on the combiner level increased from 260s to 322s, a 1.6x increase in the number of clients only leads to a 10\% increase in round time. This illustrates the nature of the map-reduce like architecture - it is important for overall throughput to optimize the ratio of clients per combiner and total number of combiners.

\begin{figure}[H]
    \centering
    \includegraphics[width=0.7\linewidth]{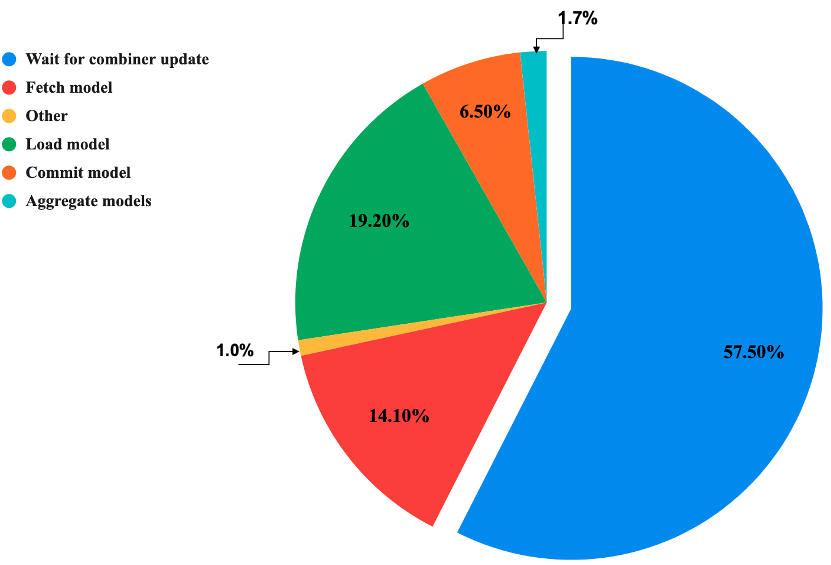}
    \caption{Workload distribution for the reducer, numbers in percentages of total round time.}
    \label{fig:redbreaks}
\end{figure}

Zooming in on the detailed workload distribution, we see that the majority of time spent at clients are related to the actual model training, with roughly 11\% overhead from downloading and sending models (Supplementary Material). Combiners spend on average only 3\% of time aggregating weight updates (Supplementary Material), the rest is spent waiting for model updates and receiving and loading weights. The reducer on the other hand spends a comparatively large fraction of time downloading and loading models from the combiners (Fig. \ref{fig:redbreaks}). The reason for this is the fact that we fetch models sequentially after all combiners finish, while the combiner implementation overlaps data transfer with computation. We plan to improve this part of the reducer in future work.

Finally we note that in each round in our final experiment, we marshall (serialization, reading and writing to disk) and transport 40GB of data between clients and the FEDn network, at an average total throughput of 465MB/s. Taken together, our experiments demonstrate that the framework is capable of sustaining the workloads that can be expected for large-scale cross-silo applications.    

\subsection{Resource utilization for a cross-device use-case}
Daily human activity recognition (DHAR) is a challenging research areas with impact on different sectors such as the health care system \cite{qi2018examining}, e.g. medical diagnosis, rehabilitation, and elderly care, and energy efficiency \cite{ ahmadi2018real}.
DHAR leverages data collected from various IoT devices (sensors, mobiles, smartwatches, etc.) that are distributed in a smart home for machine learning, with the goal to understand and learn the person's daily behaviors \cite{alkhabbas2020activity}. However, it can be privacy-invasive to share data about different users' behaviors with third-party stakeholders, hence it is a promising use-case for federated learning.

We use the CASA dataset \cite{cook2010learning} and a Long Short Term Memory (LSTM) model for human activity recognition to evaluate FEDn in a cross-device setting. The data has been collected from 30 different homes over a two month period from continuous ambient and PIR sensors distributed throughout the home, which contains $13956534$ patterns, each pattern comprises a set of $37$ features linked to different sensors, that represents the daily human activity (sleep, eat, read, watch tv,etc.) for one volunteer in each.  The model architecture comprises of one LSTM input layer, four dense layers, and one output layer, with the aim to classify the output to 10 different user daily activities. The model has in total 68,884 trainable parameters and the size of the model is $254KB$. The local training parameters setting per round are, $1$ epoch, $32$ batch size and $0.01$ learning rate with ADAM as optimizer. The same LSTM model has been used for all experiments presented in this section. 

\begin{figure}[!h]
  \centering
  \includegraphics[width=\linewidth, height=5.5cm]{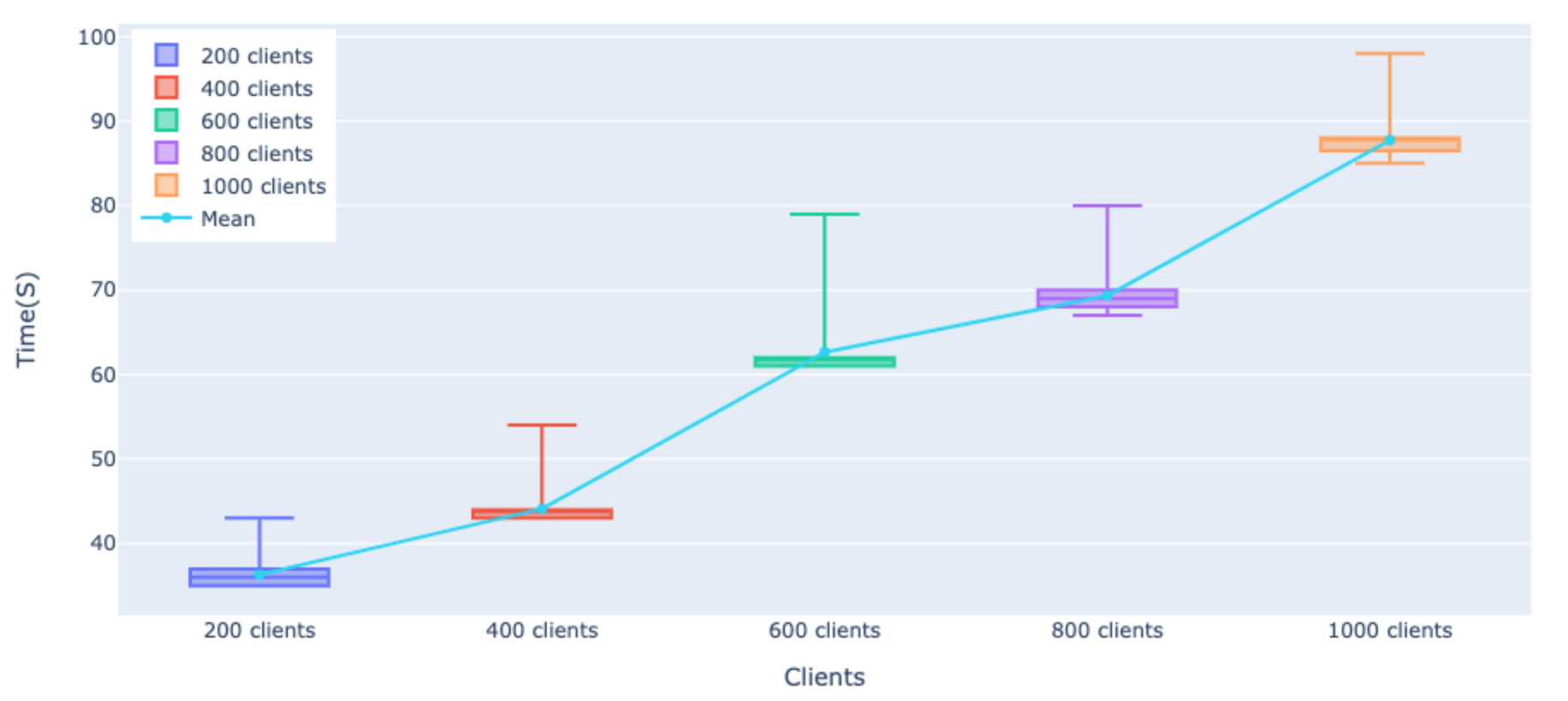}
  \caption{Round time as a function of number of clients for a single large combiner. A 5x increase in number of clients only leads to a 2x increase in round time.}
  \label{fig:combinerCapacity}
\end{figure}

Fig. \ref{fig:combinerCapacity} shows that a single large combiner can comfortably handle up to $1000$ client connections in our experiments. We studied the round time as a function of the number of clients and observed a 2x increase in round time when the number of clients grow from 200 to 1000. This illustrates the communication efficiency of the combiners. 

\subsubsection{Resource-constrained combiners}

\begin{figure*}[htp]
  \centering
  \includegraphics[width=\linewidth, height=5.5cm]{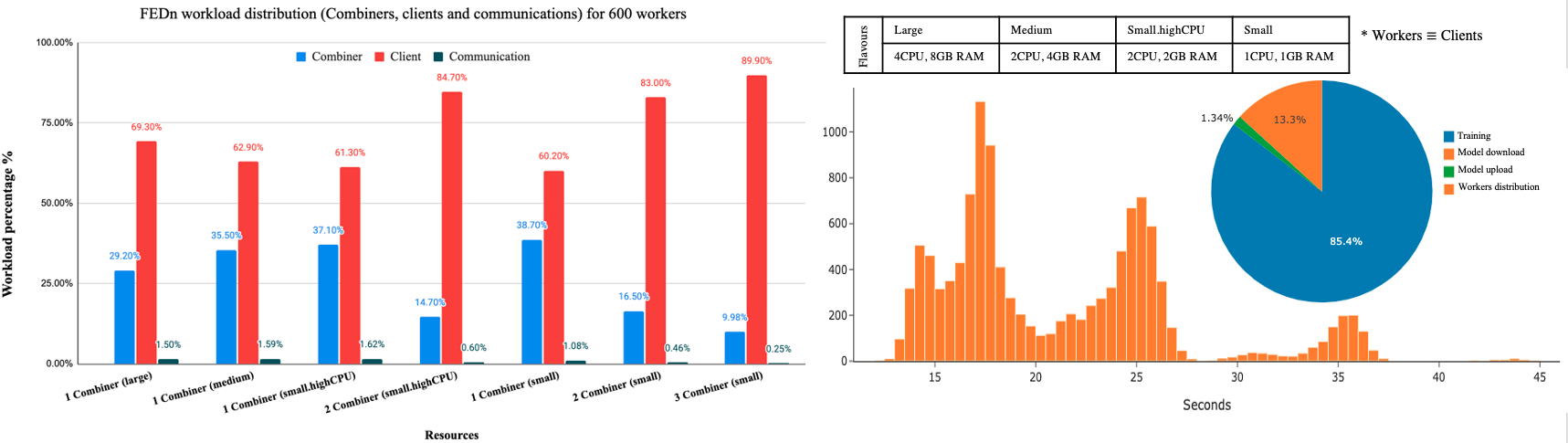}
  \caption{The left panel shows the FEDn workload distribution over combiner and clients and the communication between them for different combiner flavours and configuration. The right panel shows client training time distribution and the fraction of time used by the clients to download, train, and upload, the model respectively.}
  \label{fig:combinerFlavoursIOT}
\end{figure*}

In a fog/edge scenario it is not always possible to rely on the type of high-powered resources we used in the previous section. Here we conducted a vertical scaling study to study the overall resource cost-benefit trade-off for hosting the framework. We conducted experiments where 600 clients were allowed to connect to different combiner configurations with varying resource constraints. Figure \ref{fig:combinerFlavoursIOT} shows FEDn workload distribution over both combiner and clients while training. From the left barplot, we see that the FEDn core components do not require high-end specialized resources for operational tasks. In fact, while the percentage combiner workload using one large combiner (29.20 $\%$) is lower than with one small combiner (38.70$\%$), the FEDn core components still work reliably also in the case of a resource-constrained environment. Moreover, the figure illustrates that by replicating small combiners, the overall workload is subdivided resulting in improved performance and efficiency over a single large combiner. The experiments also illustrate that by adding more constrained-resources (small combiner), the framework performs better and better, together with increased robustness from replicated services. In fact, three small combiners perform better than a single large combiner. This result again helps to better estimate the deployment budget and also highlights the flexibility in terms of deployment strategies and effective utilization of resources of the framework.

The pie chart in Fig. \ref{fig:combinerFlavoursIOT} shows the combiner's overall workload distribution during the federated training process.  The blue color in the pie chart represents the time taken by the clients during the course of the federated training (85.4 $\%$ of the total execution time required to complete $20$ rounds). As expected, the client-side consumed most of the time in the local training and model sharing processes, highlighting the relatively low overhead from the framework federation, even using low-cost hardware for combiners. Here it is important to note that, this distribution of the workload is highly dependent on the model size, model parameters, dataset type, and data size used to train local models at each client's side (as we saw in the previous cross-silo experiments). 

The histogram chart shows the training time distribution amongst clients. We can see that a high number of clients accomplish the training process between $8-27$ seconds and a small fraction falls between $40-45$ seconds, which means that we observe some stragglers or communication delays during the execution. This illustrates that federated model training using the FEDn framework is highly efficient in its communication strategy and at the same time robust enough to manage unexpected behaviors that may happen during the training process. These features are essential to offer production-grade services in geographically distributed settings.

Cross-device federated training requires a framework that can manage high-throughput communication based on a very large number of devices, efficient processing to prepare partial and global models, and robust mechanisms to manage devices with constrained-resources (low processing or communication capabilities). The presented experiments clearly illustrate that the framework is well-suited for cross-device federated learning. 

\section{Conclusion}\label{sec:conclusion}
In this paper, we have designed and developed FEDn, a lightweight, scalable, robust, and efficient framework that is vendor-agnostic both in terms of the underlying machine learning libraries and also the distributed infrastructure technologies. In this article, we present an in-depth performance analysis based on both large numbers of clients (cross-device) and large models (cross-silo), training federated models in real, heterogeneous distributed environments. Taken together, our results highlight that FEDn can sustain federated machine learning applications from many different application areas in production environments. Furthermore, the framework is based on a structured computational model, and it is our hope that it will be a useful framework also for researchers developing new federated schemes, allowing them to rapidly test ideas in real complex distributed and heterogeneous infrastructure environments.

\section{Availability}
FEDn is publicly available under the Apache2 license at \href{https://github.com/scaleoutsystems/fedn}{https://github.com/scaleoutsystems/fedn}. The models and examples in this benchmark are available at \url{https://github.com/scaleoutsystems/examples}.

\section{Acknowledgement}
We acknowledge valuable discussions on the design with Daniel Zakrisson, Jens Frid and Stefan Hellander, and code contributions from Li Ju, Fredrik Wrede and Desislava Stoyanova. Funding has been provided by the eSSENCE strategic collaboration on eScience (Alawadi, Ait-Mlouk, Toor, and Hellander) and the Swedish Innovation Agency Vinnova grant. no. 2019-02819 (awarded to Scaleout Systems AB). 

\bibliographystyle{IEEEtranN}
\bibliography{references}

\end{document}